\documentclass[letterpaper, 10 pt, conference]{ieeeconf}

\IEEEoverridecommandlockouts                           

\usepackage{cite}
\usepackage{hyperref}
\usepackage{amsmath,amssymb,amsfonts}
\usepackage{algorithmic}
\usepackage{algorithm}
\usepackage{graphicx}
\usepackage{textcomp}
\usepackage{xcolor}
\usepackage{booktabs}
\usepackage{multirow}

\usepackage{subcaption}
\usepackage{tikz}

\usepackage{flushend} 

\title{\LARGE \bf
Scaling Vision-Language Reward Learning for Robot Manipulation in Parallel Simulation
}

\author{Lobna Joualy$^{1,3,*}$, Eric Demeester$^{2,3}$, and Nikolaos Tsiogkas$^{1,3}$%
\thanks{This work was funded by the Research Foundation of Flanders (FWO),
Grant Number 1SHHJ24N.}%
\thanks{$^{1}$Department of Computer Science and $^{2}$Department of Mechanical
Engineering, KU Leuven, 3590 Diepenbeek, Belgium. $^{3}$Flanders Make@KU Leuven,
Belgium. E-mail:
{\tt\small \{lobna.joualy,nikolaos.tsiogkas\}@kuleuven.be},
{\tt\small eric.demeester@kuleuven.be}. $^{*}$Corresponding author: Lobna Joualy.}}

\begin{document}

\raggedbottom

\maketitle
\thispagestyle{empty}
\pagestyle{empty}

%%%%%%%%%%%%%%%%%%%%%%%%%%%%%%%%%%%%%%%%%%%%%%%%%%%%%%%%%%%%%%%%%%%%%%%%%%%%%%%%
\begin{abstract}

Vision-language models (VLMs) can replace human annotators in preference-based reward learning, but sequential API requests and single-environment data collection make training slow and costly. We present RAPID (Reward learning with Adaptive Parallel Image Diversity), a system that couples GPU-parallel rollout with data-aware policy updates, single-request preference labeling, automatic reward stabilization, and representative image sampling. We evaluate these components on five Franka Panda manipulation tasks in IsaacLab. Parallel rollout and adaptive updates provide the first substantial reduction in training time: under matched two-stage prompting, mean runtime falls from 9.18 to 3.13 hours. With all RAPID components enabled, training completes in 1.15 hours using 896 rather than 19,840 API calls per run, and aggregate final success rises from 86.3\% to 98.7\%. This represents an 8.0$\times$ end-to-end speedup and a 95.5\% reduction in API usage. An offline evaluation with Gemma~3 12B and GPT-4.1 mini demonstrates that single-request prompting reduces labeling latency and cost across both models. Code is available at: \url{https://github.com/rapid-vlm/rapid-vlm-rl}.

\end{abstract}

%%%%%%%%%%%%%%%%%%%%%%%%%%%%%%%%%%%%%%%%%%%%%%%%%%%%%%%%%%%%%%%%%%%%%%%%%%%%%%%%

\section{INTRODUCTION}

Robot learning relies on reward functions that are often laborious to engineer. Preference-based learning instead trains a reward from relative judgments over pairs of robot behaviors, known as trajectory segments \cite{christiano2017deep,lee2021pebble}, and recent work shows that \emph{vision-language models} (VLMs) can provide these judgments from images without continuous human supervision \cite{wang2024rl}. Here we address a separate bottleneck in online VLM-based preference reward learning: long wall-clock training times and high VLM feedback overhead.

Modern GPU-accelerated simulators such as IsaacLab \cite{mittal2023orbit,mittal2025isaac} and IsaacGym \cite{makoviychuk2021isaac} can collect experience from many environments concurrently, but this alone does not accelerate the reward-learning process. Each parallel iteration produces more transitions, altering the balance between collected data and policy updates \cite{fedus2020revisiting}. Parallel collection may also yield similar images, while fixed feedback schedules can continue requesting VLM labels after the reward model stabilizes. We address these issues by coordinating policy updates and VLM feedback with the rate and diversity of collected experience.

We present RAPID, a parallel VLM reward-learning system built around four complementary components: (1) parallel rollout with adaptive update scaling adjusts policy updates to newly collected experience; (2) single-request VLM labeling replaces RL-VLM-F’s \cite{wang2024rl} two-stage requests with one request per image-pair comparison; (3) an adaptation of prediction-stability stopping \cite{bloodgood2009method} halts reward learning once the learned signal stabilizes; and (4) an adaptation of greedy farthest-point sampling maintains a diverse visual query pool. Together, these components form a coordinated feedback loop for high-throughput online VLM reward learning.

Our key contributions are:
\begin{enumerate}
    \item An integrated system that coordinates GPU-parallel experience collection with policy optimization, reward learning, and resource-efficient VLM feedback acquisition.
    \item A controlled study across five manipulation tasks and 135 training runs, in which RAPID achieves 98.7\% average success, reduces training time from 9.18 to 1.15 hours, and reduces VLM calls by 95.5\% relative to the RL-VLM-F baseline. Component ablations and a fixed 1,000-pair benchmark across two VLMs isolate the effects and characterize the tradeoffs of the individual design choices.
\end{enumerate}

%%%%%%%%%%%%%%%%%%%%%%%%%%%%%%%%%%%%%%%%%%%%%%%%%%%%%%%%%%%%%%%%%%%%%%%%%%%%%%%%

\section{RELATED WORK}
\label{sec:related}

\subsection{Learning Manipulation from Visual Feedback}

Robot manipulation learning has evolved from manual reward engineering to learning from demonstrations \cite{argall2009survey} and preferences \cite{wirth2017survey}. DAgger \cite{ross2011reduction}, behavioral cloning \cite{pomerleau1991efficient}, and inverse reinforcement learning \cite{abbeel2004apprenticeship} require demonstrations, while TAMER \cite{knox2009interactively} and COACH \cite{celemin2015coach} use evaluative or corrective feedback. Preference-based methods instead learn from pairwise comparisons. PEBBLE \cite{lee2021pebble} combines experience relabeling and pre-training, while SURF \cite{park2022surf} adds semi-supervised reward learning. This preference-learning formulation underlies RAPID; the following subsection reviews how VLMs can provide the needed feedback.

\subsection{Vision Language Models for Robot Learning}

Large-scale vision-language models show strong visual reasoning capabilities \cite{liu2023visual}. Foundation models have been used for robot control \cite{zitkovich2023rt}, pairwise reward feedback \cite{wang2024rl}, and absolute reward ratings \cite{luu2025enhancing}. RL-VLM-F extends PEBBLE by replacing human pairwise feedback with VLM-generated preferences over image pairs. Recent methods pursue complementary directions: RL-SaLLM-F uses trajectory augmentation and label verification \cite{tu2025online}, while PRIMT combines multimodal feedback with trajectory synthesis \cite{wang2025primt}. RAPID builds directly on RL-VLM-F’s preference-learning formulation because it provides a controlled basis for image-based reward learning from VLM feedback. By retaining its reward and policy learners, we focus on a different goal: making the learning process faster. To reduce feedback overhead, RAPID replaces the two sequential VLM requests used for each comparison with a single request. We evaluate this design through controlled training runs and an offline benchmark with two VLM families, measuring both labeling accuracy and efficiency.

\subsection{GPU-Accelerated Parallel Simulation}

The advent of GPU-accelerated physics engines has transformed robot learning. IsaacGym \cite{makoviychuk2021isaac} demonstrated that thousands of parallel environments could be simulated on a single GPU at real-time speeds, while IsaacLab \cite{mittal2023orbit, mittal2025isaac} provides a modular framework for robot learning research. These simulators enable vectorized training where multiple environment instances run synchronously in a single process \cite{freeman2021brax}, accelerating policy learning for locomotion \cite{rudin2022learning} and manipulation \cite{allshire2022transferring}.

Most parallel-simulation work assumes \textit{known rewards}. When rewards are learned online, data collection, VLM labeling, reward-model fitting, recomputing rewards for stored experience, and policy optimization become interdependent. Our contribution is not parallel simulation itself, but coordinating these stages through adaptive updates, representative selection, and reward-stability monitoring.
%%%%%%%%%%%%%%%%%%%%%%%%%%%%%%%%%%%%%%%%%%%%%%%%%%%%%%%%%%%%%%%%%%%%%%%%%%%%%%%%

\section{Background}

\subsection{Problem Formulation}

We consider manipulation tasks as Markov Decision Processes (MDPs) \textit{without predefined reward functions}: $\mathcal{M} = (\mathcal{S}, \mathcal{A}, \mathcal{T}, \gamma)$, where $\mathcal{S}$ is the state space (joint angles, object poses), $\mathcal{A}$ is the action space (joint position commands), $\mathcal{T}$ is the transition function, and $\gamma$ is the discount factor. Unlike standard RL, the reward function $\mathcal{R}$ is \textit{learned} from preference feedback rather than specified a priori.

\subsection{Preference-Based Reward Learning}

Rather than requiring a hand-crafted reward, preference-based RL learns one from human feedback \cite{christiano2017deep,lee2021pebble}. An annotator is shown pairs of short trajectory segments and indicates which one shows more progress; this comparative judgment is the \textit{preference label}. We represent each segment as $\sigma_i = \{(s_t^i, a_t^i, I_t^i)\}_{t=1}^H$, a sequence of states, actions, and rendered RGB images of length $H$. A label $\sigma_i \succ \sigma_j$ means segment $i$ is preferred. In our setting, $H=1$ (single-image segments) and labels are ternary: $0$ (segment $i$ preferred), $1$ (segment $j$ preferred) or $-1$ (no preference or uncertain), following RL-VLM-F \cite{wang2024rl}. Uncertain responses are discarded before reward-model training.

These labels are used to train a reward model $r_\theta$. Under the Bradley-Terry model \cite{bradley1952rank}, the probability that segment $i$ is preferred is:
\begin{equation}
    P[\sigma_i \succ \sigma_j] = \frac{\exp\!\left(\sum_{t=1}^H r_\theta(I_t^i)\right)}{\sum_{k \in \{i,j\}} \exp\!\left(\sum_{t=1}^H r_\theta(I_t^k)\right)}
\end{equation}
where the reward model takes rendered images as input. Minimizing the binary cross-entropy between this distribution and the collected labels trains $r_\theta$ to assign higher scores to states the labeler judged as better. We train an ensemble of $n_e=3$ such networks and use the ensemble mean as the reward signal for the RL policy.

\subsection{VLM-Based Preference Labeling}

Although human feedback can provide valuable supervision, its dependence on annotator availability limits its scalability. RL-VLM-F \cite{wang2024rl} replaces the human annotator with a vision-language model (VLM), which receives pairs of images from the replay buffer and returns a preference label automatically. It runs a \textit{two-stage} query per pair: a first API call for image analysis and a second to produce the final label. It also operates in a single environment, so even with GPU-accelerated physics the VLM query bottleneck still results in 8--10 hour training runs in IsaacLab. Section~\ref{sec:method} introduces four techniques that address both the per-label cost and the single-environment throughput bottleneck.

%%%%%%%%%%%%%%%%%%%%%%%%%%%%%%%%%%%%%%%%%%%%%%%%%%%%%%%%%%%%%%%%%%%%%%%%%%%%%%%%

\section{Method}
\label{sec:method}

\begin{figure*}[t]
    \centering
    
    % --- TOP IMAGE (a) ---
    \begin{subfigure}{\textwidth}
        \centering
        \includegraphics[width=\textwidth, trim={0cm 3.6cm 0cm 3.6cm}, clip]{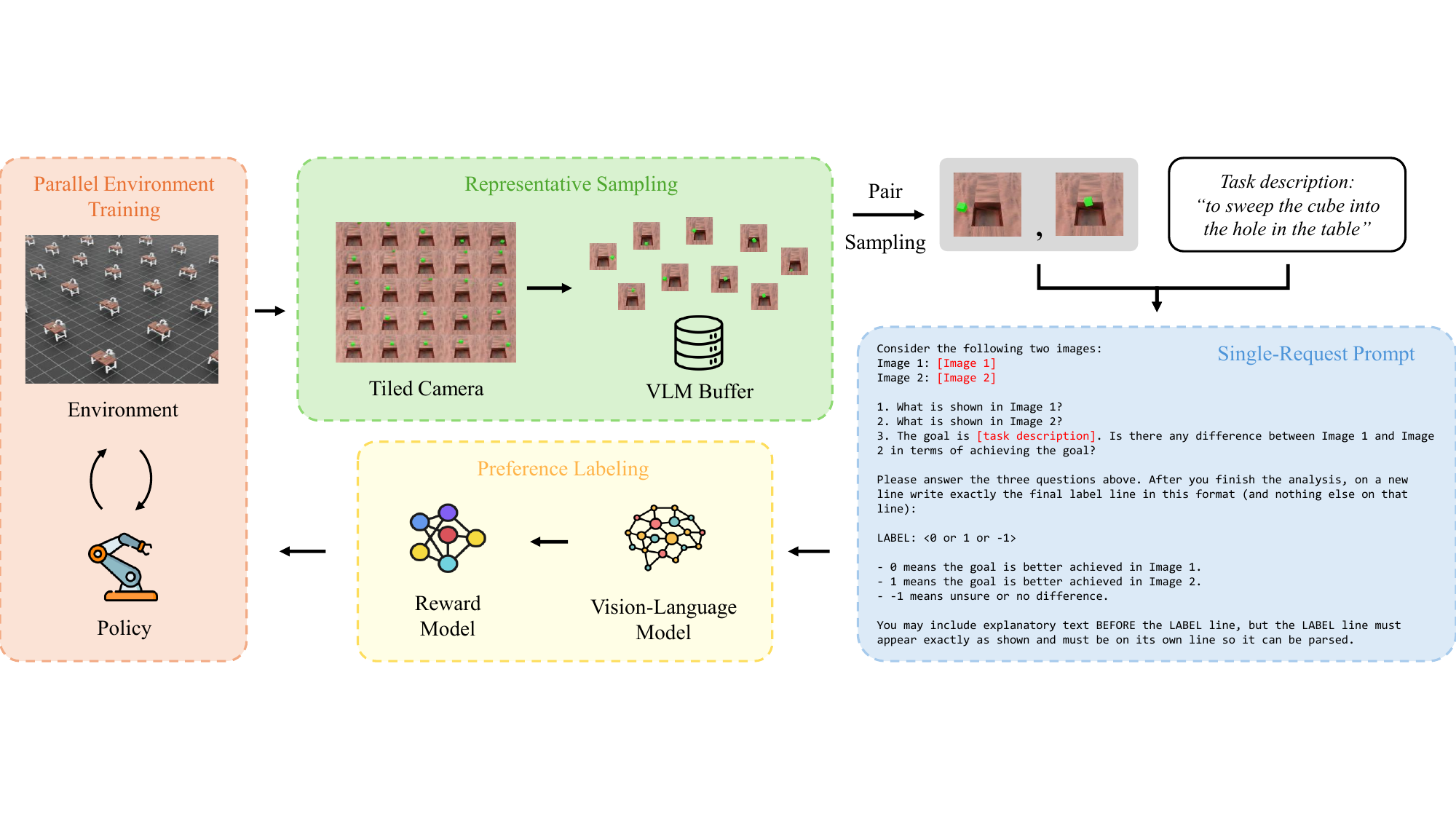}
        \caption{Active learning phase: Full preference labeling pipeline operates until reward model stabilizes.}
        \label{fig:pipeline_active}
    \end{subfigure}

    \vspace{0.25em}
    
    % --- DASHED LINE ---
    \begin{tikzpicture}
        \draw [dashed, line width=0.8pt, gray!50] (0,0) -- (0.99\textwidth,0);
    \end{tikzpicture}
    
    \vspace{0.75em}

    % --- BOTTOM IMAGE (b) ---
    \begin{subfigure}{\textwidth}
        \centering
        \includegraphics[width=\textwidth, trim={0cm 6.5cm 0cm 6.5cm}, clip]{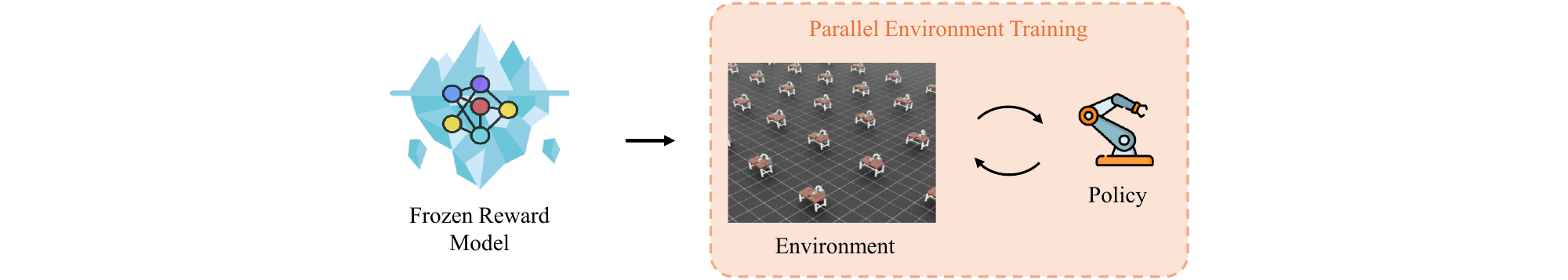}
        \caption{Frozen phase: Preference labeling pipeline stops automatically when 
        stability criteria are met ($\rho\geq0.99$, $\delta\leq0.025$, patience=2). Policy continues 
        training with frozen reward model.}
        \label{fig:pipeline_frozen}
    \end{subfigure}

    \vspace{0.5em}
    \caption{System overview. (a) Active phase: Parallel environments collect experience and render observations via a tiled camera. Representative sampling selects diverse images for the VLM buffer. Periodically, image pairs are sampled and labeled by a vision-language model using single-request queries. The reward model ensemble trains on preferences and provides learning signals to the policy. (b) Frozen phase: When correlation-based monitoring meets the fixed stability criteria, preference sampling, reward updates, and VLM queries stop. Policy training continues with the frozen reward model.}
    \label{fig:pipeline}
\end{figure*}
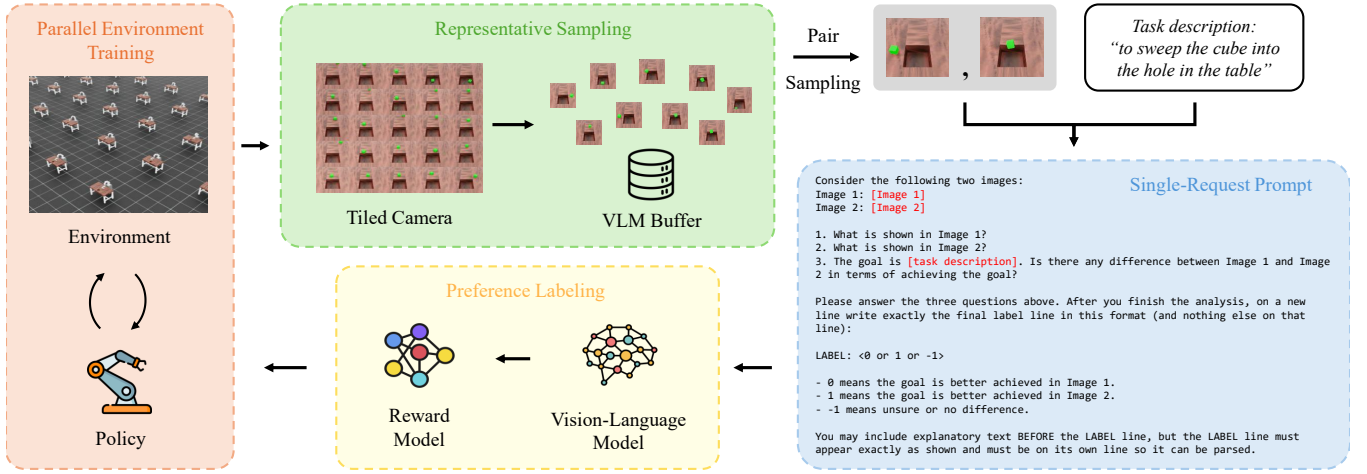
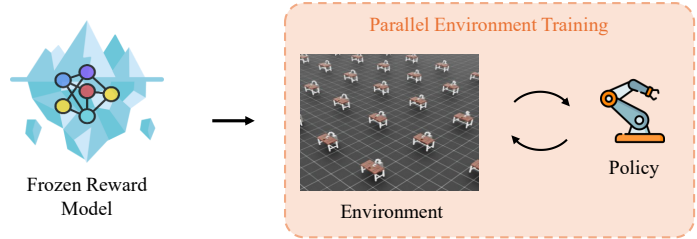

RAPID retains RL-VLM-F's preference model and policy learner, but reorganizes their interaction with data collection through four components (Figure~\ref{fig:pipeline}): (1) parallel rollout with scaled policy updates, (2) single-request VLM labeling, (3) automatic reward stabilization, and (4) representative sampling. Section~\ref{sec:results} measures the combined gain from parallel rollout and its scaled update schedule, followed by the gains from the remaining components.

\subsection{Parallel Environment Training with Adaptive Updates}

In the single-environment implementation, each simulator iteration produces one transition and schedules Soft Actor-Critic (SAC) updates around it. With \(E_{\text{active}}\) parallel environments, each iteration instead produces \(E_{\text{active}}\) transitions. Keeping a fixed update count per iteration would therefore change the ratio of policy updates to collected data. We allocate updates by the number of new transitions and cap optimization bursts after initial data collection to limit primacy bias in the off-policy networks \cite{nikishin2022primacy}.

Following standard multi-environment off-policy practice \cite{stable-baselines3}, we maintain a fractional \textit{update budget}. Let $t$ be the budgeted update iterations, $\Delta N_t$ the number of transitions collected since the previous iteration, $\tilde b_t$ the available budget before executing updates, $n_t$ the number of SAC updates executed, and $b_t$ the residual credit carried to the next iteration. The implemented schedule is
\begin{align}
    \tilde b_t &= \min\!\left(b_{t-1}+\alpha\Delta N_t,\,U_{\max}+1\right), \\
    n_t &= \left\lfloor\min\!\left(\tilde b_t,\,U_{\max}\right)\right\rfloor, \\
    b_t &= \tilde b_t-n_t .
\end{align}
Here $\alpha$ is the target number of SAC updates per transition and $U_{\max}$ limits a single optimization burst. We selected \(\alpha=0.5\) in preliminary experiments and fixed it across all runs; with 50 active environments, this allocates 25 updates per collection iteration. We chose \(U_{\max}=50\) so that the cap remains above the steady-state budget but prevents large accumulated bursts, particularly immediately after the seed phase. Logs show that the cap is reached only on the first budgeted iteration after the 10K-step seed phase, leaving one unit of credit. The next iteration executes 26 updates and subsequent iterations execute 25.

\subsection{Single-Request VLM Labeling}

The baseline two-stage prompting uses two sequential API calls per preference decision: the first receives the image pair and returns a free-form analysis, and the second maps that analysis to label 0, 1, or $-1$. This doubles the request count and increases latency.

The \textit{single-request} variant (Figure~\ref{fig:pipeline_active}) sends both images, the task description, and the required output schema in one VLM request. It therefore halves API calls per image-pair comparison. We cap output length and parse the same ternary label set as the baseline. The objective is lower latency and cost. We examine the effect of removing the intermediate analysis on label accuracy in Section~\ref{sec:prompt_results}.

\subsection{Automatic Reward Stabilization}
\label{sec:reward_stabilization}

Reward models can drift during extended training, particularly as the policy distribution shifts under parallel collection. A fixed query budget does not adapt to convergence: it can stop too early on a difficult run or continue querying after consecutive reward models have become effectively unchanged.

The idea behind \textit{correlation-based stability monitoring} is simple: a converged reward model should assign consistent scores to the same states across consecutive updates. Inspired by Bloodgood and Vijay-Shanker \cite{bloodgood2009method}, who propose halting labeling when model predictions on a held-out set stop changing across rounds, we maintain an \textit{anchor set} $\mathcal{A}$ of $N$ images sampled from the replay buffer and check consistency after each reward model update:
\begin{enumerate}
    \item Compute predictions: $\vec{r}_t = \{r_\theta(I) : I \in \mathcal{A}\}$
    \item If $t > 0$, compute metrics:
    \begin{align}
        \rho_t &= \text{corr}(\vec{r}_{t-1}, \vec{r}_t) \quad \text{(Pearson)} \\
        \delta_t &= \frac{1}{N} \sum_{i=1}^N |r_t^{(i)} - r_{t-1}^{(i)}|
    \end{align}
    \item If $\rho_t \geq \rho_{\text{thresh}}$ AND $\delta_t \leq \delta_{\text{thresh}}$:
    \begin{align*}
        c_{\text{stable}} &\leftarrow c_{\text{stable}} + 1
    \end{align*}
    \quad else: $c_{\text{stable}} \leftarrow 0$
    \item If $c_{\text{stable}} \geq P$: FREEZE (stop reward updates and VLM queries)
    \item Store: $\vec{r}_{t-1} \leftarrow \vec{r}_t$
\end{enumerate}

We selected \(\rho_{\text{thresh}}=0.99\), \(\delta_{\text{thresh}}=0.025\), and \(P=2\) in small-scale pilot experiments over neighboring values, as they yielded the most consistent stopping across all five tasks. The correlation threshold requires near-perfect linear agreement, the absolute-change threshold limits shifts in reward magnitude, and the patience criterion prevents freezing after a single transient measurement. We then fixed these values for every reported seed without task-specific tuning.

If the correlation and delta conditions are not met, reward learning and VLM querying continue until the feedback budget or training horizon is reached.

When freezing is triggered, the system transitions from the active phase 
(Figure \ref{fig:pipeline_active}) to the frozen phase (Figure \ref{fig:pipeline_frozen}), 
where the entire preference labeling pipeline stops, while policy training continues.

\subsection{Representative Sampling for Diverse Preferences}

FIFO sampling from the replay buffer can select visually similar trajectories, particularly when synchronized environments visit nearby states. Such comparisons add little coverage while consuming VLM calls.

We adapt \textit{greedy farthest-point sampling} \cite{gonzalez1985clustering, sener2018active}. Before each labeling round, $M$ recent candidate images are converted to L2-normalized grayscale feature vectors. Starting from the middle candidate, each iteration adds the image with the greatest minimum L2 distance to the selected set, until $K$ images are retained. Pairs are sampled from this refreshed pool. We use fixed image features because they are inexpensive and independent of the reward model; reward-encoder features change as the reward model is updated and would couple query selection to the model whose convergence is being monitored. RAPID uses 16$\times$16 grayscale features; Section~\ref{sec:feature_results} compares this choice with 32$\times$32 and 64$\times$64 pixels and with reward-encoder features.

Algorithm~\ref{alg:parallel_rollout} summarizes how the four components interact in the complete training loop.

\begin{algorithm}[t]
\caption{RAPID Training Loop}
\label{alg:parallel_rollout}
\footnotesize
\begin{algorithmic}[1]
\STATE Initialize policy $\pi$, reward ensemble $\{r_\theta^{(k)}\}_{k=1}^{n_e}$
\STATE Initialize $\mathcal{D}$; $b,N,N_{\mathrm{prev}},c_{\mathrm{stable}}\leftarrow0$; $\mathrm{labeled},\mathrm{frozen}\leftarrow\mathrm{false}$
\WHILE{$N < T_{\text{total}}$}
    \STATE Collect and store transitions and images; $N \leftarrow N+E_{\text{active}}$
    \IF{$N\geq N_{\mathrm{seed}}$ AND NOT $\mathrm{labeled}$}
        \STATE Select representative pool; obtain initial VLM labels
        \STATE Relabel $\mathcal{D}$ with current ensemble; reset critic; perform initial SAC update
        \STATE Fix stability anchor $\mathcal{A}$; $\mathrm{labeled}\leftarrow\mathrm{true}$
    \ELSIF{$\mathrm{labeled}$}
        \STATE $\Delta N_t \leftarrow N-N_{\mathrm{prev}}$; $b \leftarrow \min(b+\alpha\Delta N_t,U_{\max}+1)$
        \STATE $n \leftarrow \lfloor\min(b,U_{\max})\rfloor$
        \FOR{$i = 1$ to $n$}
            \STATE Sample batch from $\mathcal{D}$; perform one SAC update
        \ENDFOR
        \STATE $b \leftarrow b-n$; $N_{\mathrm{prev}}\leftarrow N$
    \ENDIF
    \IF{reward update due AND feedback remains AND NOT $\mathrm{frozen}$}
        \STATE Refresh representative pool; query VLM; fit $\{r_\theta^{(k)}\}$; relabel $\mathcal{D}$
        \STATE Compute $(\rho_t,\delta_t)$ on $\mathcal{A}$; update $c_{\mathrm{stable}}$
        \IF{$c_{\mathrm{stable}}\geq P$}
            \STATE $\mathrm{frozen}\leftarrow\mathrm{true}$; stop reward updates and VLM queries
        \ENDIF
    \ENDIF
\ENDWHILE
\end{algorithmic}
\end{algorithm}

%%%%%%%%%%%%%%%%%%%%%%%%%%%%%%%%%%%%%%%%%%%%%%%%%%%%%%%%%%%%%%%%%%%%%%%%%%%%%%%%

\section{Experimental Evaluation}

\subsection{Manipulation Tasks}

\begin{figure*}[t]
    \centering
    \begin{minipage}{2.65cm}
        \centering
        \begin{tikzpicture}
            \begin{scope}
                \clip (-1.325cm,-1.325cm) rectangle (1.325cm,1.325cm);
                \node at (0,0) {\includegraphics[height=2.65cm]{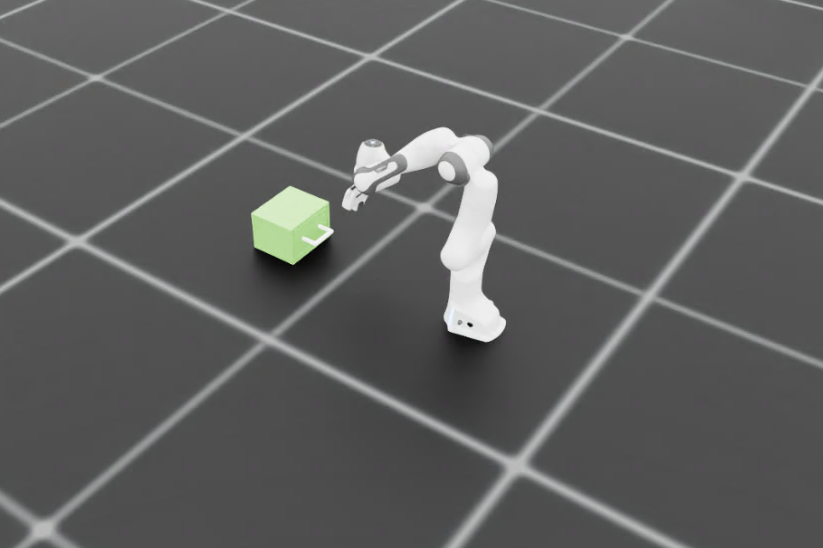}};
            \end{scope}
            \useasboundingbox (-1.325cm,-1.325cm) rectangle (1.325cm,1.325cm);
        \end{tikzpicture} \\
        \small{Open Drawer}
    \end{minipage}
    \hfill
    \begin{minipage}{2.65cm}
        \centering
        \begin{tikzpicture}
            \begin{scope}
                \clip (-1.325cm,-1.325cm) rectangle (1.325cm,1.325cm);
                \node at (0,0) {\includegraphics[height=2.65cm]{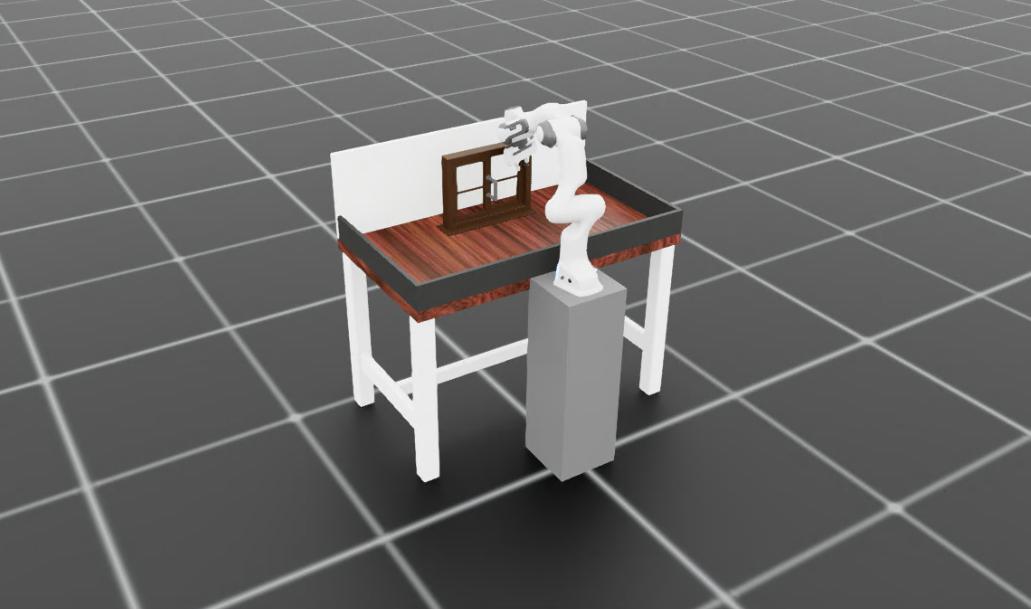}};
            \end{scope}
            \useasboundingbox (-1.325cm,-1.325cm) rectangle (1.325cm,1.325cm);
        \end{tikzpicture} \\
        \small{Open Window}
    \end{minipage}
    \hfill
    \begin{minipage}{2.65cm}
        \centering
        \begin{tikzpicture}
            \begin{scope}
                \clip (-1.325cm,-1.325cm) rectangle (1.325cm,1.325cm);
                \node at (0,0) {\includegraphics[height=2.65cm]{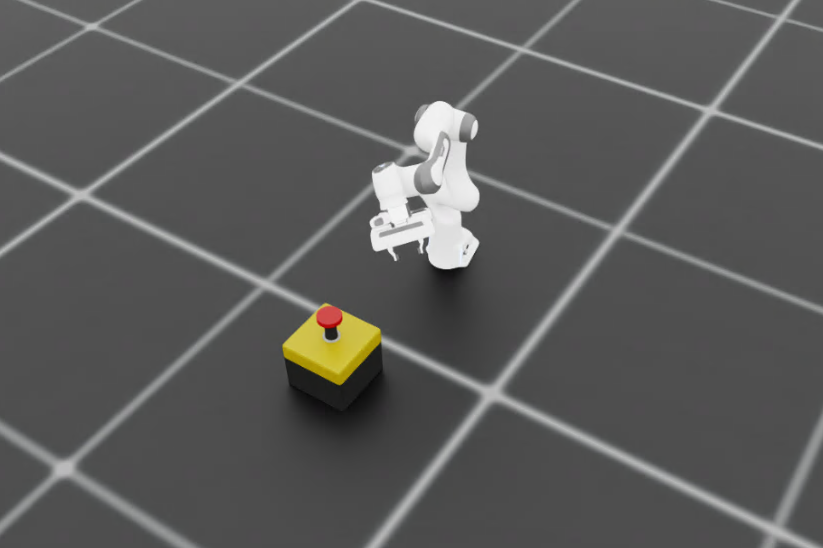}};
            \end{scope}
            \useasboundingbox (-1.325cm,-1.325cm) rectangle (1.325cm,1.325cm);
        \end{tikzpicture} \\
        \small{Push Button}
    \end{minipage}
    \hfill
    \begin{minipage}{2.65cm}
        \centering
        \begin{tikzpicture}
            \begin{scope}
                \clip (-1.325cm,-1.325cm) rectangle (1.325cm,1.325cm);
                \node at (0,0) {\includegraphics[height=2.65cm]{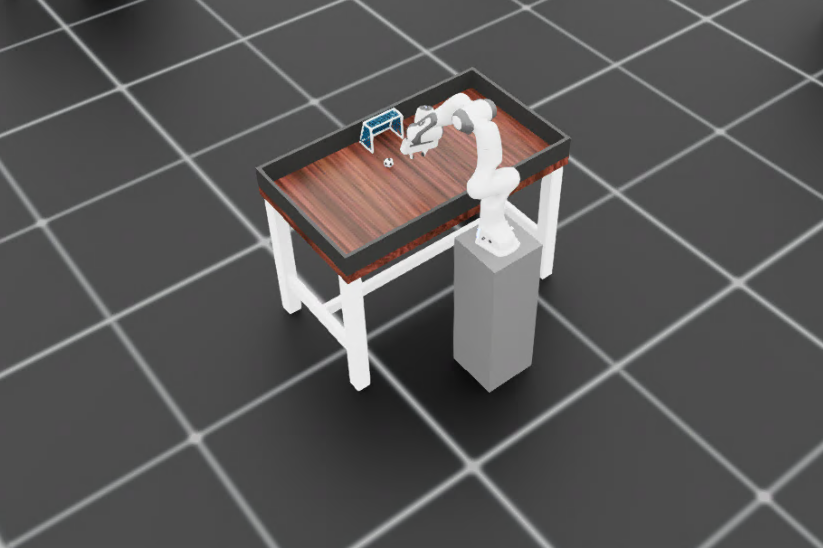}};
            \end{scope}
            \useasboundingbox (-1.325cm,-1.325cm) rectangle (1.325cm,1.325cm);
        \end{tikzpicture} \\
        \small{Soccer}
    \end{minipage}
    \hfill
    \begin{minipage}{2.65cm}
        \centering
        \begin{tikzpicture}
            \begin{scope}
                \clip (-1.325cm,-1.325cm) rectangle (1.325cm,1.325cm);
                \node at (0,0) {\includegraphics[height=2.65cm]{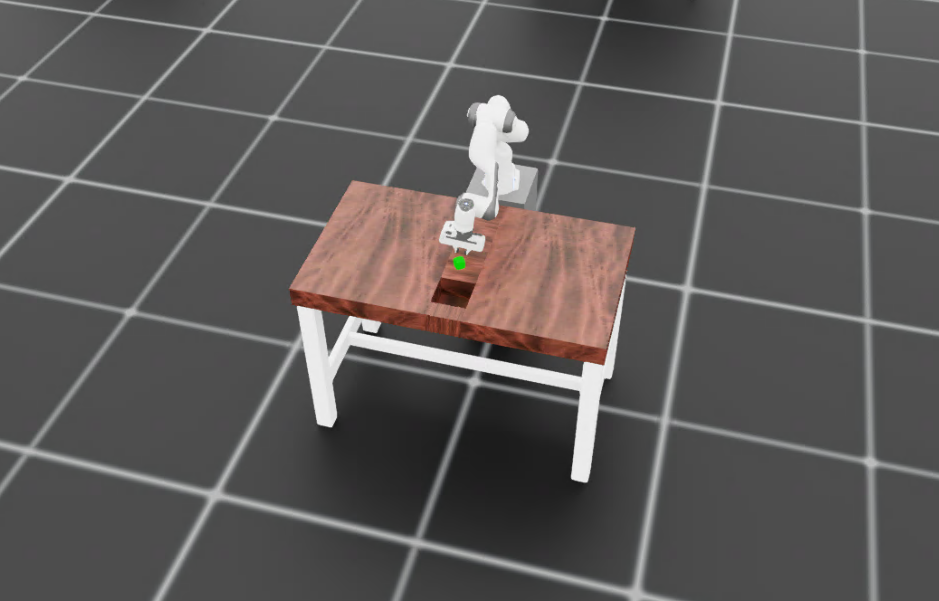}};
            \end{scope}
            \useasboundingbox (-1.325cm,-1.325cm) rectangle (1.325cm,1.325cm);
        \end{tikzpicture} \\
        \small{Sweep Into}
    \end{minipage}

    \vspace{6pt}
    \caption{Manipulation task environments. Left to right: Open Drawer, Open Window, Push Button (top-down), Soccer (ball-into-goal), Sweep Into (cube-into-hole).}
    \label{fig:environments}
\end{figure*}
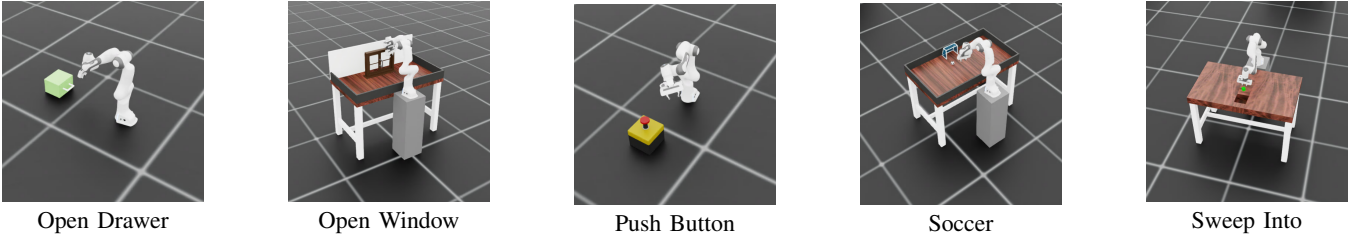

We evaluate on five manipulation tasks derived from MetaWorld \cite{yu2020meta} and adapted to IsaacLab with Franka (Figure \ref{fig:environments}):

\begin{itemize}
    \item \textit{Open Drawer:} Grasp a drawer handle and pull outward.
    \item \textit{Open Window:} Slide a window open along a horizontal track.
    \item \textit{Push Button:} Press a button downward.
    \item \textit{Soccer:} Push a soccer ball into a goal net.
    \item \textit{Sweep Into:} Sweep a cube into a table hole.
\end{itemize}

The policy's reward signal is supplied by the preference reward model; task-success signals are used only for evaluation and to construct the offline oracle benchmark.

\subsection{Ablation Study Design}

The primary ablation adds components in the order used by the completed runs:

\begin{itemize}
    \item \textbf{C1 (Baseline):} our IsaacLab implementation of RL-VLM-F, with one environment, two-stage prompting, FIFO sampling, and no stability stopping.
    \item \textbf{C2 (+Single Request):} C1 with one VLM request per pair.
    \item \textbf{C3 (+Parallel):} C2 with 50 environments and adaptive updates ($\alpha=0.5$, $U_{\text{max}}=50$).
    \item \textbf{C4 (+Stability):} C3 with correlation-based stopping ($\rho \geq 0.99$, $\delta \leq 0.025$, $P=2$).
    \item \textbf{C5 (RAPID):} C4 with representative sampling ($K=50$, $M=5000$, 16$\times$16 grayscale features).
\end{itemize}

To isolate the effect of prompt format from collection regime, we additionally run a 50-environment variant with two-stage prompting and the C3 update schedule. Together with C1--C3, this forms a $2\times2$ comparison over prompt format and collection regime (one-environment baseline vs 50-environment adaptive). Because environment count and update schedule change jointly, the comparison measures their combined effect. We run three seeds for every task and condition: 75 primary ablation and 15 prompt-comparison runs. Each run uses 1M environment steps and evaluates 10 episodes every 10K steps.

For each seed, ``final success'' is the mean of the last 10 evaluations (last 100K steps). Per-task tables report the mean and sample standard deviation over three seeds; aggregate results report variation over the 15 task--seed summaries. We report wall-clock duration and API calls, with sample standard deviation summarizing variation across seeds.

\textbf{Baseline choice.} We use our single-environment IsaacLab implementation of RL-VLM-F (C1), rather than timing the original MuJoCo-based MetaWorld results. IsaacLab and MuJoCo \cite{todorov2012mujoco} differ in physics, contact modeling, rendering, and hardware utilization. By evaluating all conditions in the same simulator, codebase, hardware, and training budget, we avoid cross-simulator timing confounds and attribute observed differences to the controlled configuration changes.

\subsection{VLM and Sampling Benchmarks}

The ablations do not reveal whether prompt performance generalizes across VLMs or whether representative sampling performance depends on the chosen image representation. We therefore examine these questions in two controlled studies.

\textbf{Prompt-format study.} We fix 1,000 rollout image pairs and vary only the VLM and prompt format. The benchmark contains 200 pairs per task (150 clear and 50 deliberately ambiguous). A scripted oracle prioritizes success, then task progress, and marks near-equal pairs as ambiguous. To assess the effect of prompt structure, Gemma~3 12B and GPT-4.1 mini evaluate identical pairs and task instructions using one request or two sequential requests. Accuracy is measured on the 750 clear pairs; latency and cost use all pairs, the same OpenRouter backend, and the 40-request training concurrency. This isolates the model-dependent accuracy--efficiency tradeoff of prompt format.

\textbf{Sampling-representation study.} We hold C5 fixed and vary only the features used for farthest-point selection: 16$\times$16, 32$\times$32, or 64$\times$64 grayscale pixels, or the current reward-encoder features (6,144 concatenated dimensions). Success, runtime, and API calls reveal whether the sampling result is specific to the 16$\times$16 choice. Each representation is evaluated across 15 task–seed runs. The \(16\times16\) condition reuses C5, and the three alternatives contribute 45 additional runs, for 135 unique training runs overall.

\subsection{Implementation Details}

\textbf{Simulation.} We use IsaacLab 2.2.0 on a RTX 5090 GPU (32GB VRAM).

\textbf{Observation \& Action.} We define the policy input as a state including robot joint positions and velocities, task-relevant object states, and the last action. Fixed-viewpoint 128$\times$128 RGB images are used only for VLM queries. Actions are 7-dim joint position with 1-dim binary gripper command (8-dim total).

\textbf{Camera Setup.} A fixed third-person TiledCamera renders all 50 environment views simultaneously at 128$\times$128 resolution; the tiled output is split into per-environment frames.

\textbf{Policy Learning.} We use SAC \cite{haarnoja2018soft} with the RL-VLM-F actor and critic: hidden dimension 256, depth 3, and ReLU activations. Actor and critic learning rates are $3\times10^{-4}$ and batch size is 512. Training lasts 1M environment steps, including an initial 10K-step random seed phase.

\textbf{Reward Model.} Following PEBBLE \cite{lee2021pebble} and RL-VLM-F \cite{wang2024rl}, we use an ensemble of three convolutional networks to reduce variance from noisy preference labels at moderate compute cost. Each has four convolutional layers (16, 32, 64, 128 channels; kernel sizes 5, 3, 3, 3) and a fully connected output, trained with Adam at $3\times10^{-4}$.

\textbf{VLM.} Policy-training runs use Gemma~3 12B \cite{gemmateam2025gemma3technicalreport} through OpenRouter. Prompt structure is fixed within each prompt condition. GPT-4.1 mini \cite{openai2025gpt41} is used only in the offline benchmark.

\textbf{Reward Stabilization.} We use the correlation-and-delta criterion in Section~\ref{sec:reward_stabilization} with the same thresholds for every task and seed.

\textbf{VLM Preference Collection.} Every 4,000 environment steps, the system sends 40 image-pair comparisons to the VLM. C5 samples from 50 representative images selected from up to 5,000 recent candidates; earlier configurations use FIFO sampling. The stability anchor contains $N=512$ images.

\begin{figure*}[t]
    \centering
    \includegraphics[width=\textwidth]{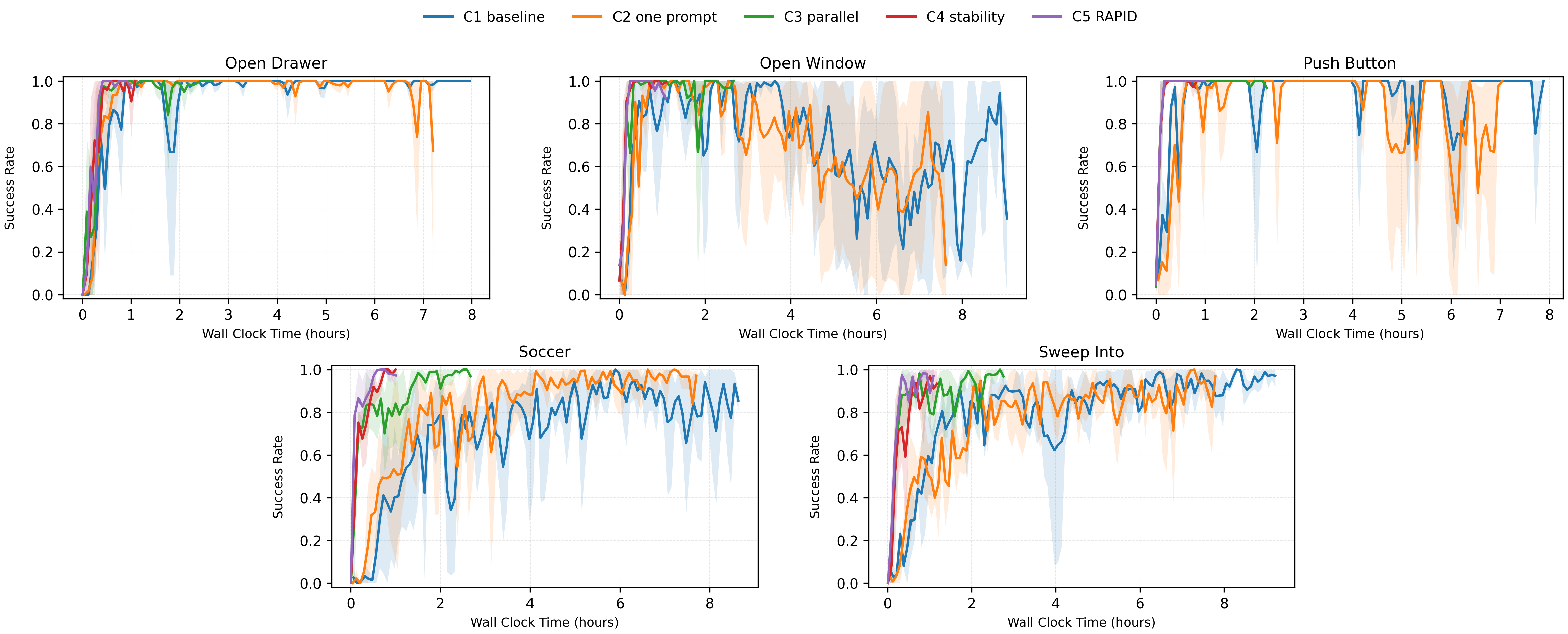}
    \caption{Success versus wall-clock time (mean $\pm$ sample standard deviation, three seeds and linear interpolation at five-minute intervals over the shared observed time range).}
    \label{fig:learning_curves}
\end{figure*}

%%%%%%%%%%%%%%%%%%%%%%%%%%%%%%%%%%%%%%%%%%%%%%%%%%%%%%%%%%%%%%%%%%%%%%%%%%%%%%%%

\section{Results}
\label{sec:results}

\subsection{Wall-Clock and Task Performance}

For the same 1M environment steps, mean duration falls from 9.18 to 1.15 hours, an 8.0$\times$ end-to-end speedup; per-task C1 means span 8.45--9.60 hours and C5 means span 1.07--1.23 hours. With two-stage prompting fixed, the combined change to 50-environment rollout and adaptive updates reduces time from 9.18 to 3.13 hours (2.9$\times$). Within the 50-environment single-request pipeline, stability stopping and representative sampling reduce time from 2.74 hours (C3) to 1.15 hours (C5), a further 2.4$\times$ reduction from RAPID's query-management components.

Figure~\ref{fig:learning_curves} reports mean success with sample-standard-deviation bands over three seeds. Table~\ref{tab:success_rates} gives the step-controlled result: C5 averages 98.7\% over the last 100K steps, compared with 86.3\% for C1. The per-task means and standard deviations quantify variation across the three seeds.

The difficult C1 tasks are Open Window and Soccer. C5 reaches 95--100\% on every task. With 50 parallel environments, RAPID achieves a measured 8.0$\times$ end-to-end speedup, reflecting the remaining costs of policy optimization, rendering, and VLM queries.

\begin{table}[t]
\centering
\caption{Final success (\%, mean$\pm$sample standard deviation over three seeds), where each seed is averaged over its last 10 evaluations.}
\label{tab:success_rates}
\resizebox{\columnwidth}{!}{%
\begin{tabular}{lccccc}
\toprule
\textbf{Task} & \textbf{C1} & \textbf{C2} & \textbf{C3} & \textbf{C4} & \textbf{C5} \\
\midrule
Open Drawer   & \textbf{100.0$\pm$0.0} & 93.3$\pm$5.8 & 96.3$\pm$6.4 & 98.7$\pm$0.6 & 95.0$\pm$3.6 \\
Open Window   & 65.3$\pm$11.0 & 44.7$\pm$27.9 & 95.3$\pm$8.1 & 96.7$\pm$3.5 & \textbf{100.0$\pm$0.0} \\
Push Button   & 84.7$\pm$18.6 & 57.0$\pm$25.2 & 99.3$\pm$1.2 & 99.3$\pm$1.2 & \textbf{100.0$\pm$0.0} \\
Soccer        & 84.7$\pm$18.0 & 97.0$\pm$2.6 & 99.0$\pm$1.0 & 98.7$\pm$1.2 & \textbf{100.0$\pm$0.0} \\
Sweep Into    & 96.7$\pm$1.5 & 93.0$\pm$5.0 & 98.0$\pm$1.7 & 95.3$\pm$3.8 & \textbf{98.3$\pm$0.6} \\
\midrule
\textbf{Aggregate} & 86.3$\pm$16.5 & 77.0$\pm$26.8 & 97.6$\pm$4.3 & 97.7$\pm$2.6 & \textbf{98.7$\pm$2.4} \\
\bottomrule
\end{tabular}
}
\end{table}

\subsection{Component Ablation and Prompt-Format Comparison}
\label{sec:prompt_results}

Table~\ref{tab:config_summary} summarizes the sequential C1--C5 ablation and includes the additional 50-environment variant with two-stage prompting used for the $2\times2$ prompt/collection comparison. Moving from C1 to C2 halves API calls and reduces time by 16\%, but decreases aggregate success by 9.3 percentage points. Moving from C2 to C3 adds parallel rollout and adaptive updates, reducing time by 2.8$\times$ and increasing aggregate success by 20.6 points. Moving from C3 to C4 cuts calls by 81.0\% and time by 2.1$\times$ with a 0.1-point success change. Finally, moving from C4 to C5 cuts a further 52.5\% of calls and adds 1.0 point of success.

\begin{table*}[t]
\centering
\caption{Component ablation and prompt-format comparison. C1--C5 form the sequential ablation; the final row is the additional 50-environment variant with two-stage prompting. Success, time, and API calls are mean$\pm$sample standard deviation over 15 task--seed runs.}
\label{tab:config_summary}
\begin{tabular}{lccccccl}
\toprule
\textbf{Config.} & \textbf{Envs} & \textbf{Prompt} & \textbf{Stability} & \textbf{Representative} & \textbf{Success (\%)} & \textbf{Time (h)} & \textbf{API calls} \\
\midrule
C1 & 1  & two & no  & no  & 86.3$\pm$16.5 & 9.18$\pm$0.73 & 19,840$\pm$0 \\
C2 & 1  & one & no  & no  & 77.0$\pm$26.8 & 7.75$\pm$0.36 & 9,920$\pm$0 \\
C3 & 50 & one & no  & no  & 97.6$\pm$4.3  & 2.74$\pm$0.21 & 9,920$\pm$0 \\
C4 & 50 & one & yes & no  & 97.7$\pm$2.6  & 1.31$\pm$0.47 & 1,888$\pm$2,537 \\
C5 & 50 & one & yes & yes & \textbf{98.7$\pm$2.4} & \textbf{1.15$\pm$0.08} & \textbf{896$\pm$656} \\
\midrule
50-env. two-stage & 50 & two & no & no & 87.1$\pm$25.7 & 3.13$\pm$0.27 & 19,840$\pm$0 \\
\bottomrule
\end{tabular}
\end{table*}

Together, C1, C2, C3, and the 50-environment variant with two-stage prompting in Table~\ref{tab:config_summary} form a $2\times2$ comparison of prompt format (one versus two stages) and collection regime (one versus 50 environments). With one environment, C1 with two-stage prompting attains higher final success than single-request C2 (86.3\% versus 77.0\%). At 50 environments, single-stage C3 attains higher and less variable success than the two-stage variant (97.6\% versus 87.1\%). For the 50-environment two-stage variant, Open Window reaches $78.0\pm38.1\%$ and Push Button reaches $63.3\pm40.4\%$, with variability reported as sample standard deviation across three seeds. The preferred prompt format therefore changes with the collection regime, so downstream policy success cannot be attributed to VLM label accuracy alone.

\begin{figure}[H]
    \centering
    \includegraphics[width=\columnwidth]{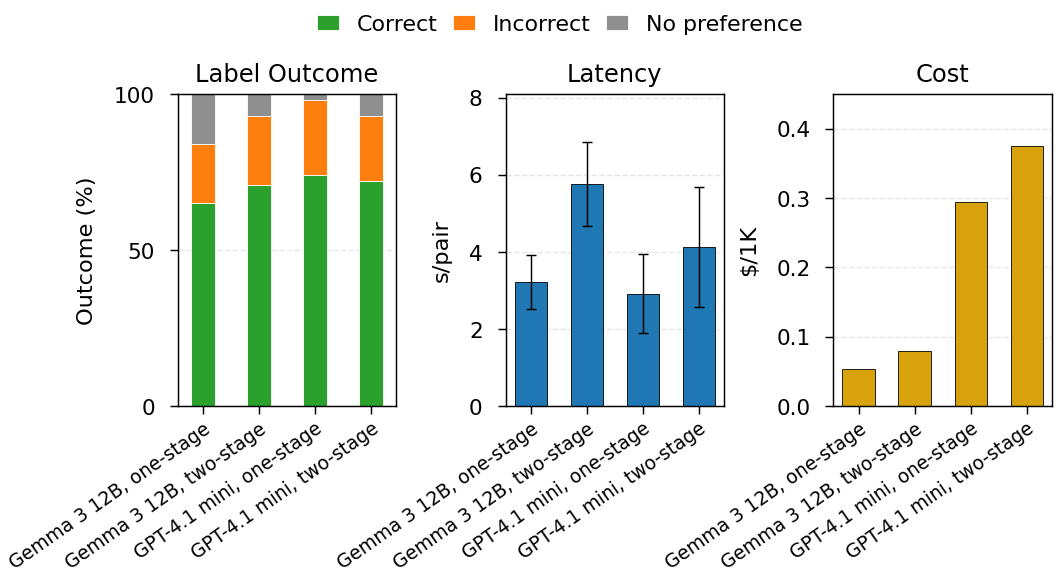}
    \caption{Offline comparison of Gemma 3 12B and GPT-4.1 mini under single-request and sequential prompting. Outcomes use 750 clear pairs, while latency and cost cover the full 1,000-pair benchmark.}
    \label{fig:vlm_benchmark}
    \vspace{-12pt}
\end{figure}

Across both VLM families in Figure~\ref{fig:vlm_benchmark}, single-request prompting reduces latency, cost, and API calls. GPT-4.1 mini performs better with single-request prompting (74.0\% versus 72.0\% accuracy). Gemma favors two-stage prompting (70.9\% versus 65.1\%), revealing an explicit accuracy--efficiency tradeoff: its single-request format is 32.4\% cheaper, 44.2\% faster, and uses half as many API calls per image-pair comparison. Gemma provides the lowest observed labeling cost in this benchmark; its single-request format costs approximately 5.5$\times$ less per comparison than single-request GPT-4.1 mini. Thus model and prompt choice can be matched to the available labeling budget and required oracle accuracy.

\begin{table}[t]
\centering
\caption{Sampling-feature ablation; each row summarizes 15 task--seed runs. Success is mean$\pm$sample standard deviation; other columns are means.}
\label{tab:feature_resolution}
\footnotesize
\setlength{\tabcolsep}{5pt}
\begin{tabular}{lccc}
\toprule
\textbf{Representation} & \textbf{Success (\%)} & \textbf{Time (h)} & \textbf{API calls} \\
\midrule
16$\times$16 pixels & \textbf{98.7$\pm$2.4} & \textbf{1.15} & \textbf{896} \\
32$\times$32 pixels & 96.0$\pm$6.5 & 1.36 & 2,373 \\
64$\times$64 pixels & 87.9$\pm$20.6 & 1.53 & 3,128 \\
Reward encoder & 89.7$\pm$15.2 & 1.65 & 3,771 \\
\bottomrule
\end{tabular}
\end{table}

\subsection{Reward Stability}
\label{sec:stability_results}

The stability criterion stops querying in all 15 C5 runs at 22K--274K steps (mean 95.6K), and in 14 of 15 C4 runs at 26K--506K. C5 averages 896 calls, 91.0\% fewer than C3 and 95.5\% fewer than C1, while retaining 98.7\% final-window success. The sole non-stopping C4 run continues to the training horizon.

\begin{figure}[t]
    \vspace{-6pt}
    \centering
    \includegraphics[width=0.92\columnwidth]{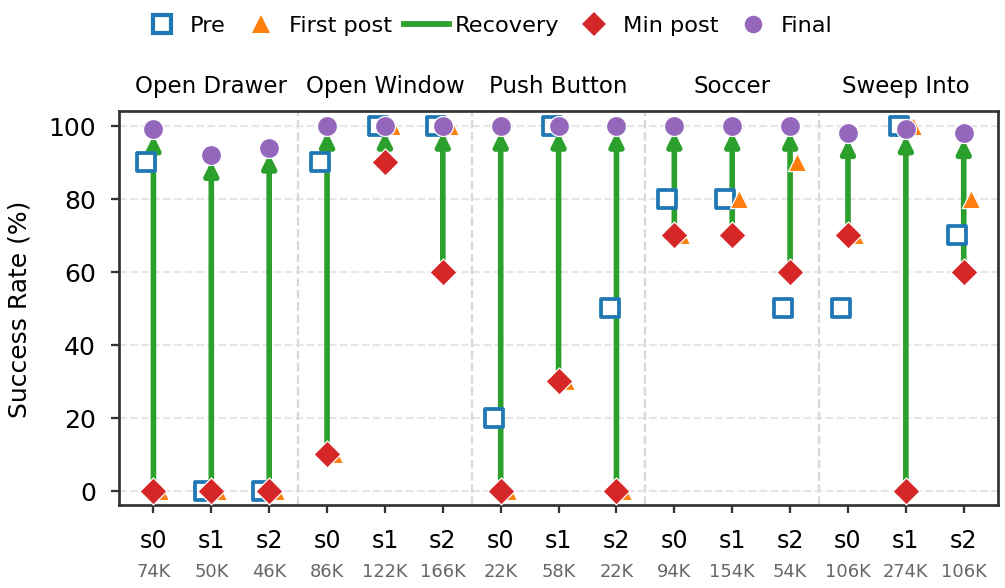}
    \caption{Success before and after reward freezing across 15 C5 task--seed runs. The logged freeze step is shown below each run. With evaluations every 10K steps, Pre and First post denote the evaluations immediately before and after freezing, while Min post denotes the lowest subsequent success. Arrows show recovery from Min post to the mean of the final 10 evaluations.}
    \label{fig:freeze_final}
    \vspace{-16pt}
\end{figure}

Figure~\ref{fig:freeze_final} shows the nearest evaluations before and after each freeze event, with policy performance measured every 10K steps. Eleven of the 15 runs show a transient decline after freezing, but all subsequently reach 92--100\% success in the final evaluation window.

\subsubsection{Sampling Representation}
\label{sec:feature_results}
Table~\ref{tab:feature_resolution} compares fixed pixels with reward-encoder features. At 16$\times$16, fixed pixels reach 98.7\% success in 1.15 hours with 896 calls, versus 89.7\%, 1.65 hours, and 3,771 calls for the encoder; stopping occurs in 15 of 15 versus 10 of 15 runs. These results support fixed low-resolution pixels for these tasks.

%%%%%%%%%%%%%%%%%%%%%%%%%%%%%%%%%%%%%%%%%%%%%%%%%%%%%%%%%%%%%%%%%%%%%%%%%%%%%%%%

\section{Discussion}

\subsection{Impact on Robot Learning Workflows}

RAPID changes the experimental cycle from an overnight run to about 69 minutes on our hardware. Parallel rollout with adaptive updates supplies the first 2.9$\times$ reduction; single-request labeling, adaptive stopping, and representative selection make the feedback loop inexpensive enough to realize an 8.0$\times$ end-to-end improvement. At the observed averages, 50 C5 runs require about 44,800 API calls rather than 992,000 for C1.

\subsection{Applicability and Design Choices}

Single-request labeling improves efficiency, while its effect on accuracy is model-dependent: offline accuracy increases for GPT-4.1 mini and decreases for Gemma. Gemma remains the less expensive option, with a per-comparison cost approximately 5.5$\times$ lower.
Stability stopping provides the largest query reduction, stopping feedback in every C5 run and 14 of 15 C4 runs. Representative sampling lowers the mean and variability of call count, while the matched study supports fixed 16$\times$16 features over changing reward-encoder features.

\subsection{Limitations and Future Directions}

The study is entirely in simulation and does not demonstrate sim-to-real transfer. The reward model observes rendered images from one fixed camera, while the policy receives simulator state; lighting, texture, viewpoint, dynamics, and contact differences may therefore invalidate the learned reward on a physical robot. Domain randomization, real-image calibration, and physical evaluation are required before making a transfer claim \cite{allshire2022transferring,rudin2022learning}. Real-robot-only learning also cannot obtain the parallel-physics speedup.

Other limitations include the five short-horizon tasks and testing the second VLM only offline. We also test the current reward encoder but not fixed pretrained embeddings, which may better capture subtle semantics. Future work may examine such representations and feedback reactivation under distribution shift. RAPID could also be extended to long-horizon tasks through an LLM-based task planner that decomposes an objective into subgoals, with RAPID learning the corresponding sub-rewards without manual reward specification.

%%%%%%%%%%%%%%%%%%%%%%%%%%%%%%%%%%%%%%%%%%%%%%%%%%%%%%%%%%%%%%%%%%%%%%%%%%%%%%%%

\section{Conclusion}

We presented RAPID, an integrated system for VLM preference reward learning in GPU-parallel simulation. Parallel rollout and adaptive updates provide a 2.9$\times$ combined gain; reward stabilization and representative sampling further reduce runtime from 2.74 to 1.15 hours and calls from 9,920 to 896. Across five manipulation tasks, RAPID reaches 98.7\% success versus 86.3\% for the baseline, yielding an 8.0$\times$ end-to-end speedup and 95.5\% fewer calls. The controlled studies expose model-dependent prompt tradeoffs and post-freeze recovery, providing a basis for future sim-to-real work.

%%%%%%%%%%%%%%%%%%%%%%%%%%%%%%%%%%%%%%%%%%%%%%%%%%%%%%%%%%%%%%%%%%%%%%%%%%%%%%%%

\bibliographystyle{IEEEtran}
\bibliography{bibliography}

\end{document}